\documentclass{article}

\usepackage{arxiv}

\usepackage[utf8]{inputenc} 
\usepackage[T1]{fontenc}    
\usepackage{hyperref}       
\usepackage{url}            
\usepackage{booktabs}       
\usepackage{amsfonts}       
\usepackage{nicefrac}       
\usepackage{microtype}      

\usepackage{lipsum}         
\usepackage{graphicx}
\usepackage{natbib}
\usepackage{doi}
\usepackage{makecell}
\usepackage{amssymb,cuted}
\usepackage{mathtools} 
\usepackage{amsmath} 
\usepackage{breqn}
\usepackage{multirow}
\usepackage{adjustbox}
\usepackage{cleveref}       
\graphicspath{{./images/}} 
\crefdefaultlabelformat{#2\bfseries\upshape(#1)#3}
\creflabelformat{eq.}{#2\bfseries\upshape(#1)#3}

\title{Global Pointer: Novel Efficient Span-based Approach for Named Entity Recognition}

\author{
	Jianlin Su \\
	Zhuiyi Technology Co., Ltd.\\
	Shenzhen, China \\
	\texttt{bojonesu@wezhuiyi.com} \\
	\And
	Ahmed Murtadha \\
	Zhuiyi Technology Co., Ltd.\\
	Shenzhen, China \\
	\texttt{mengjiayi@wezhuiyi.com} \\
	\And
	Shengfeng Pan \\
	Zhuiyi Technology Co., Ltd.\\
	Shenzhen, China \\
	\texttt{nickpan@wezhuiyi.com} \\
		\And
	Jing Hou\\
	School of Economics and Management,\\ University of Chinese Academy of Sciences\\
	Beijing, China \\
	\texttt{houjing21@mails.ucas.ac.cn} \\
	\And
	  Jun Sun\\
	Zhuiyi Technology Co., Ltd.\\
	Shenzhen, China \\
	\texttt{jainasun@wezhuiyi.com} \\
	\And
	  Wanwei Huang\\
	Zhuiyi Technology Co., Ltd.\\
	Shenzhen, China \\
	\texttt{huangwanwei@wezhuiyi.com} \\
	\And
	Bo Wen \\
	Zhuiyi Technology Co., Ltd.\\
	Shenzhen, China \\
	\texttt{brucewen@wezhuiyi.com} \\
	\And
	Yunfeng Liu \\
	Zhuiyi Technology Co., Ltd.\\
	Shenzhen, China \\
	\texttt{glenliu@wezhuiyi.com} \\
}
\renewcommand{\headeright}{}
\renewcommand{\undertitle}{}
\renewcommand{\shorttitle}{Global Pointer}

\hypersetup{
	pdftitle={Global Pointer}
}

\begin{document}
	\maketitle
	
	\begin{abstract}
	Named entity recognition (NER) task aims at identifying entities from a piece of text that belong to predefined semantic types such as person, location, organization, etc. 
	The state-of-the-art solutions for flat entities NER commonly suffer from  capturing the fine-grained semantic information in underlying texts.  The existing span-based approaches overcome this limitation, but the computation time is still a concern.  
	In this work, we propose a novel span-based NER framework, namely Global Pointer (GP), that leverages the relative positions through a multiplicative attention mechanism. 
	The ultimate goal is to enable a global view that considers the beginning and the end positions to predict the entity. To this end, we design two modules to identify the head and the tail of a given entity to enable the inconsistency between the training and inference processes. Moreover, we introduce a novel classification loss function to address the imbalance label problem. In terms of parameters, we introduce a simple but effective approximate method to reduce the training parameters.
	We extensively evaluate GP on various benchmark datasets. Our extensive experiments demonstrate that GP can outperform the existing solution. Moreover, the experimental results show the efficacy of the introduced loss function compared to softmax and entropy alternatives.
	\end{abstract}


	\section{introduction}\label{introduction}
	
	Named entity recognition (NER) task aims to recognize entities, also called mentions, from a piece of text that belong to predefined semantic types such as person, location, organization, etc. 
	NER is a key component in natural language processing (NLP) systems for information retrieval, automatic text summarization, question answering, machine translation, knowledge base construction, etc.\cite{guo2009named,petkova2007proximity,aone1999trainable,molla2006named,babych2003improving,etzioni2005unsupervised}. Note that NER has been introduced in two forms, including flat and nested entities. 
	Flat NER has been widely addressed as a sequence labeling problem \cite{lample_etal_2016_neural}. 
	Nested entities have shown importance in various real-world applications due to their multi-granularity semantic meaning \cite{alex2007recognising,yuan2020unsupervised}. 
	However, a given token may have multiple labels and thus renders applying sequence labeling-based approaches unattainable \cite{finkel2009nested}.

	With the rapid development of deep neural network (DNN), NER task has experienced a shift towards the contextual representation learning. 
	The earlier DNN-based approaches have treated NER as a sequence labeling problem \cite{1508.01991,wang-etal-2020-pyramid,lample-etal-2016-neural}. They commonly attempt to address each token individually by capturing the type and position information.  Despite the effectiveness of these approaches, they cannot perform span-based NER, also called nested NER, in which the entity consists of more than one token \cite{finkel2009nested}. 
	DNN-based approaches for nested NER usually attempt to learn span-specific deep representation in order to classify the  corresponding type\cite{zheng2019boundary,wadden-etal-2019-entity,tan2020boundary,wang-etal-2020-pyramid,yu-etal-2020-named}.
	Recently, nested NER has experienced a shift towards  pretrained language model. Several works show that the fine-tuning approach for span representation and classification can achieve satisfactory results \cite{luan-etal-2019-general,zhong2020frustratingly}.
	The authors of \cite{yuan2021fusing} introduced modeling heterogeneous factors (e.g., inside tokens) to enhance span representation learning.

	Despite the effectiveness of the aforementioned approaches for nested NER,  the representation of a given span is simply the combination of its head and tail and thus ignores the boundary information. 
	To carry out a segment classification, the number of segments is set to the maximum length of span. Moreover, the low-quality spans, especially with long entities, dominate the corpus and thus requires high computational costs. 
	To address the aforementioned limitations, there exist some approaches initiated the solution. The authors of \cite{FuHuang-347} proposed to take the span length information into account during the training process.  Another work \cite{ShenMa-346} introduced to jointly address span classification and boundary regression in a unified framework to alleviate boundary information issue. However, the implantation of these approaches is a bit complicated and may be bothersome in real-world scenarios.

	In this paper, we propose a novel solution, namely Global Pointer (GP), to address span-based NER task. Specifically, we leverage the relative positions through a multiplicative attention mechanism  \cite{2104.09864}. 
	The ultimate goal is to enable a global view that considers the beginning and the end positions (i.e., the head and tail information) to predict the entity. To achieve this, we design two modules to identify the head and the tail of a given entity to enable the inconsistency between the training and inference processes.
	In addition, to alleviate the burden of class imbalance in NER, we extend the softmax and cross-entropy in a universal loss function. It is noteworthy that the number of parameters of the proposed solution increases when a new entity type is added. Note that the introduced loss can be applied to any task suffering from the label imbalance issue. To remedy this issue, we introduce another extension of  GP, namely efficient GP, based on an effective approximate method to reduce the number of parameters. 
	We extensively evaluate GP on various benchmark datasets. Our extensive experiments demonstrate that GP can outperform the existing solution. Moreover, the experimental results show the efficacy of the introduced loss function compared to softmax and entropy alternatives.   
	
	In brief, the main contributions are three-fold: 
	\begin{itemize}
		\item We propose a novel solution, namely Global Pointer (GP), to address span-based NER task that leverages the relative positions through a multiplicative attention mechanism.
		\item we extend the softmax and cross-entropy in a universal loss function to perform class imbalance scenarios, NER is an example. In addition, we propose an effective approximation method to reduce the training parameters when a new entity type is added.
		\item We extensively evaluate the proposed solution on various benchmark datasets. Our extensive experiments demonstrate that the proposed solution can outperform the existing solutions. Moreover, the experimental results validate the efficacy of the introduced loss function compared to softmax and entropy alternatives.   	
	\end{itemize}
	
	The remaining of the paper is organized as follows. Section \ref{sec:related_work} reviews related work. Section \ref{sec:approach} describes the propose solution. Section \ref{sec:experiments} presents the experimental settings and empirically evaluates the performance of the proposed solution. Finally, we conclude this paper with Section \ref{sec:conclusion}.

	\section{Related work}\label{sec:related_work}
	NER has received extensive attention of researchers in the last decades. The earlier solutions include rule-based \cite{kim2000rule,sekine2004definition,hanisch2005prominer,quimbaya2016named}, Unsupervised learning \cite{etzioni2005unsupervised,zhang2013unsupervised},  Feature-based supervised learning approaches \cite{szarvas2006multilingual,liu2011recognizing,rocktaschel2012chemspot}. However, the performance of these approaches heavily relies on feature extraction and hand-crafted rules, which may be bothersome in real-world scenarios. 
	
	With the rapid development of deep neural networks, various approaches were introduced to address NER task as a classification problem \cite{zhang2015fixed}. The key idea is to learn entity-specific representation to model the semantic relation between two entities.
	Convolutional neural networks \cite{yao2015biomedical,strubell2017fast,zhai2017neural}, recursive neural networks \cite{li2017leveraging,gridach2017character,wang2018code,akbik2018contextual,liu2019towards,ghaddar2018robust} and long-short term memory based approaches \cite{1508.01991,tran2017named,jie2019dependency}. The authors of  \cite{zheng2017joint,zhou2017joint} introduced to jointly extract the entities and their relations in a unified framework.
	
	Recently, pre-trained language models (PLMs) have mostly achieved the state-of-the-art performance of various NLP tasks \cite{devlin2018bert,liu2019roberta,yang2019xlnet}. Following this approach, NER has experienced a shift towards PLMs. An end-to-end model based on sequence-to-sequence learning with copy mechanism and the graph convolutional networks, which introduced to jointly extract  relation and entity from sentences \cite{zeng2018extracting,fu-etal-2019-graphrel}. A reinforcement learning-based approach  \cite{zeng-etal-2019-learning} was proposed to tackle  the extraction order of relation extraction task. A cascade binary tagging-based framework \cite{wei-etal-2020-novel} was introduced to treat relations as functions mapping subjects to objects in a sentence to alleviate the overlapping problem in relation extraction. Table-Sequence \cite{wang2020two} consists of two encoders, including a table encoder and a sequence encoder, that work together to learn  the entity-specific representation. A partition filter network-based approach \cite{YanZhang-351} introduced to model two-way interaction between  entity and relation extraction tasks. The authors of \cite{yuan2021fusing} introduced modeling relevant features by leveraging  heterogeneous factors, e.g.,  inside tokens, boundaries, and related spans to enhance learn span representation, resulting in accurate classification performance.

	\section{Approach}\label{sec:approach}
	\begin{figure*}  	
		\centering 
		\includegraphics[scale=0.5]{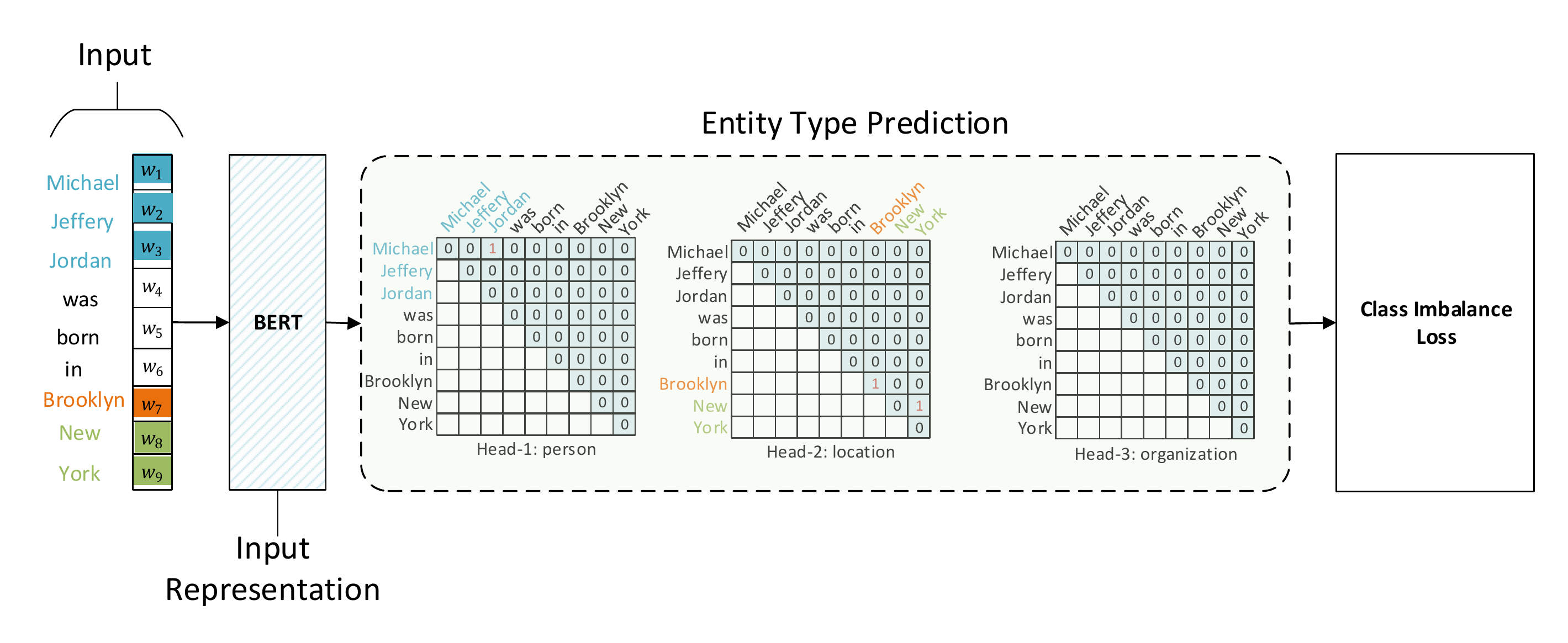} 
		\caption{An example of our proposed Global Pointer.}  
		\label{fig:env}  
	\end{figure*} 
	In this section, we describe the proposed solution. We begin by defining  span-based NER task. Then, we present the technical details of our approach. Finally, we present the approximation method to reduce the number of parameters.
	
	\subsection{Problem definition}
	Named Entity Recognition (NER) task aims to extract the entity segments and then correspondingly identify their types in the given text. Let $ S=[s_{1},s_{2},...s_{m}] $ be the possible spans in the sentence. The span $s$ is represented as $ s[i:j] $ where $i$ and $j$ are the head and tail indexes, respectively. The goal of NER is to identify all $s\in E$, where $E$ is the entity type set.
	
	\subsection{Global Pointer}
	The architecture of our proposed GP  consists of two layers, including token representation span prediction. An illustrative example of GP is shown in Figure \ref{fig:env}.

	\subsubsection{Token Representation}
	Given a sentence $ X=[x_{1},x_{2},...x_{n}] $ with $n$ token, we begin by associating each token in $X$ with its corresponding representation in the pre-training language model (PLM), e.g., BERT. We end up with a new matrix $H\in \mathbb{R}^{n\times v}$, where $v$ is dimension of representation: 
	
	\begin{equation}
		h_{1},h_{2},...h_{n}=PLM(x_{1},x_{2},...x_{n}).
	\end{equation}
	
	\subsubsection{Span Prediction}
	Now that we have already obtained the sentence representation $H$, we then compute the span representation. To this end, we use two feedforward layers that rely on the begin and end indices of the span.
	
	\begin{equation}
		{q}_{i,\alpha}={W}_{q,\alpha}{h}_i+{b}_{q,\alpha},
	\end{equation} 
	
	\begin{equation}
		{k}_{i,\alpha}={W}_{k,\alpha}{h}_i+{b}_{k,\alpha},
	\end{equation}  
	where $ {q}_{i,\alpha} \in\mathbb{R}^{d} $, $ {k}_{i,\alpha} \in\mathbb{R}^{d} $ is the vector representation of the token which used to identify the entity of type $ \alpha $. Specifically, the representation of the start and end position is ${{q}_{i,\alpha}} $ and $ {{k}_{i,\alpha}} $  for span $ s[i:j] $ of type $ \alpha $. 
	Then, the score of the span $ s[i:j] $ to be an entity of type $\alpha$ is calculated as follows:
	\begin{equation}	
		s_{\alpha}(i,j) = {q}_{i,\alpha}^{\top}{k}_{j,\alpha} 
	\end{equation} 	 
	
	
	To leverage the boundary information, we explicitly inject relative position information to the model. We apply ROPE position coding into the entity representation, which satisfies $ {\mathcal{R}}_i^{\top}{\mathcal{R}}_j ={\mathcal{R}}_{j-i} $. In this way, our scoring function is calculated as follows:
	
	\begin{dmath}	
		s_{\alpha}(i,j) = ({\mathcal{R}}_i{q}_{i,\alpha})^{\top}({\mathcal{R}}_j{k}_{j,\alpha}) = {q}_{i,\alpha}^{\top} {\mathcal{R}}_i^{\top}{\mathcal{R}}_j{k}_{j,\alpha} = {q}_{i,\alpha}^{\top} {\mathcal{R}}_{j-i}{k}_{j,\alpha} 
	\end{dmath}	 
	
	\subsection{Parameter Reduction}\label{sec:reduce_param}
	It is noteworthy to mention that when ${W}_{q,\alpha},{W}_{k,\alpha}\in\mathbb{R}^{v\times d}$, the parameters increase to $ 2vd $ for each new added entity type. Compared with the method of sequence labeling, the increase of parameters under the same conditions is about $2v$. 
	Generally speaking, $v>>d$, in the bert-base model $v$ is 768, while the common choice of d is 64.
	
	To alleviate this issue, we introduce an approximation technique to enable Global Pointer to perform under fewer parameters settings.  In the next sections, we refer to it as Efficient Global Pointer. The key idea is to capture the shared score calculation under each entity type. Specifically, we treat NER task as two subtasks, including extraction and classification. The former extracts segments as entities, and the latter identifies the type of each entity. In this way, the extraction step is equivalent to the NER task with only one entity type. We can complete it with a scoring matrix  
	$({W}_q{h}_i)^{\top}({W}_k{h}_j) $. 
	The classification step can be read as 
	$ {w}_{\alpha}^{\top}[{h}_i;{h}_j] $,
	where $ {w}_{\alpha}\in\mathbb{R}^{2v} $ denotes the identification of the entity type $ \alpha $, and $ [{h}_i;{h}_j] $ is the span representation, which is  the concatenation of the start and end representations .
	The new scoring function is the combination of :
	
	\begin{equation}	
		s_{\alpha}(i,j) = ({W}_q{h}_i)^{\top}({W}_k{h}_j) + {w}_{\alpha}^{\top}[{h}_i;{h}_j].
	\end{equation}
	
	Note that the extraction task's parameters are shared by all entity types. Therefore, when a new entity type is added, the parameters of classification task increase by $2v$, which is less compared to the original number of parameters $2vd$.
	
	To further reduce the parameters, we consider using $ [q_{i};k_{i}] $ instead of $ h_{i} $ to represent a token. Then, the final scoring function becomes:
	\begin{equation}
		s_{\alpha}(i,j) = {q}_i^{\top}{k}_j + {w}_{\alpha}^{\top}[{q}_i;{k}_i;{q}_j;{k}_j],
	\end{equation}
	where $ {w}_{\alpha} \in\mathbb{R}^{4d} $,
	$ [{q}_i;{k}_i;{q}_j;{k}_j] $ is the span representation.
	Intuitively, the number of parameters increases for each new entity type is $4d$, which is indeed less than  And $4v$.
	
	\subsection{Class Imbalance Loss}
	Inspired by the circle loss, we introduce a loss function to alleviate class imbalance.
	In single-class classification, the cross-entropy loss function is:
	
	\begin{dmath}
		\log \frac{e^{s_t}}{\sum\limits_{i=1}^n e^{s_i}}=-\log \frac{1}{\sum\limits_{i=1}^n e^{s_i-s_t}}=\log \sum\limits_{i=1}^n e^{s_i-s_t}=\log \left(1 + \sum\limits_{i=1,i\neq t}^n e^{s_i-s_t}\right),
	\end{dmath}
	
	
	where $ s_i $ is the non target score and $ s_t $ is the target score. Here, we consider the loss function in the scenario of multi-label classification. The goal is to make the score of the target class not less than that of the non-target class. Therefore, the loss function is:
	
	\begin{equation}
		\log \left(1 + \sum\limits_{i\in\Omega_{neg}} e^{s_i}\sum\limits_{j\in\Omega_{pos}} e^{-s_j}\right)	
	\end{equation}
	
	where $ \Omega_{pos} $ and $ \Omega_{neg} $ are positive sample set and negative sample set, respectively. Considering the multi-label scenario where the number of classes is not fixed, we introduce an additional class $ TH $ as the threshold value. We expect that the scores of target classes are greater than $ s_{TH} $ and those of non-target classes are less than $ s_{TH} $. Then, the loss function is calculated as:
	\begin{equation}\label{eq:loss}
		\log \left(1 + \sum\limits_{i\in\Omega_{neg},j\in\Omega_{pos}} e^{s_i-s_j}+\sum\limits_{ i\in\Omega_{neg}} e^{s_i-s_{TH}} +\sum\limits_{j\in\Omega_{pos}} e^{s_{TH}-s_j}\right)
	\end{equation}
	Equation \ref{eq:loss} can be further simplified as follows:
	\begin{equation}
		\log \left(e^{s_{TH}} + \sum\limits_{i\in\Omega_{neg}} e^{s_i}\right) + \log \left(e^{-s_{TH}} + \sum\limits_{j\in\Omega_{pos}} e^{-s_j}\right) 
	\end{equation}
	
	For sake of simplicity, we set the threshold to 0 and the final loss function:
	\begin{equation}	
		log \left(1 + \sum\limits_{i\in\Omega_{neg}} e^{s_i}\right) + \log \left(1 + \sum\limits_{j\in\Omega_{pos}} e^{-s_j}\right)\label{eq:final} 
	\end{equation}
	
	Specifically, the entity type of $ \alpha $ is represented by:
	\begin{equation}	
		\log \left(1 + \sum\limits_{(q,k)\in P_{\alpha}} e^{-s_{\alpha}(q,k)}\right) + \log \left(1 + \sum\limits_{(q,k)\in Q_{\alpha}} e^{s_{\alpha}(q,k)}\right)
	\end{equation}
	where $ q $, $ k $ represent the start and tail indexes of a span, $ P_{\alpha} $ represents a collection of spans with entity type $ \alpha $, $ Q_{\alpha} $ represents a collection of spans that are not entities or whose entity type is not $ \alpha $, $ {s_{\alpha}(q,k)} $ is the score that a span $ s[q:k] $ is an entity of type $ \alpha $.
	
	In inference step, the segments that satisfy $ s_{\alpha}(q,k)>0 $ are the output of the entity of type $ \alpha $.
	
	\section{Experiments and Evaluation }\label{sec:experiments}
	
\begin{table}
	\centering
	\begin{tabular}{lccccc}
		\hline
		Dataset       		&Train	&Test& Sentence length& Number of Entities \\
		\hline
		The People's daily 	&23,182	&46,36& 46.93 &3    \\
		CLUENER     		&10,748	&1,343& 37.38 &  10   \\
		CMeEE     			&15,000	&5,000& 54.15 & 9  \\
		CONLL04     		&4,270	&1,079& 28.77 & 4  \\
		Genia     			&16,692	&1,854& 25.35 &5 \\
		NYT 				&56,195	&5,000& 128 &-\\
		WebNLG 				&5,019	&703& 128&-\\
		ADE 				&\multicolumn{2}{c}{4,272 (10-fold)}&128& 2\\
		\hline		
	\end{tabular}
	\label{tab:Statistics}
	\caption{Statistics of datasets.}
\end{table}

\subsection{Experimental Setup }
	\textbf{Dataset}.
	To validate the proposed solution, we conduct extensive experiments on various benchmark datasets. Specifically, we rely on three Chinese NER datasets, including The People's daily, CLUENER \cite{xu2020cluener2020} and  CMeEE \cite{hongying2020building}, which has been widely used in the literature. Moreover, we also experiment with various English datasets, including CONLL04 \cite{roth2004linear}, Genia \cite{ohta2002genia}, NYT \cite{riedel2010modeling}, WebNLG \cite{zeng2018extracting} and ADE \cite{gurulingappa2012development}. Note that CMeEE and Genia were designed for nested NER task, while the others are flat task. Table \ref{tab:Statistics} shows the statistics of the datasets.
	
	\textbf{Evaluation Metrics}.
	We use strict evaluation metrics that if the entity type and the corresponding entity boundary are correct, the entity is correct. We use F1-score
	to evaluate the performance of our model.

	\textbf{Parameter Settings}.
	We use 12 heads and layers and keep the dropout probability at 0.1 with 30 epochs. The initial learning rate is $2e-5$ for all layers with a batch size of 32
	Note that we used the bert-base model \cite{devlin2018bert} to initialize the weights of our GP with Adam optimizer.
	
	\textbf{Comparative Baselines}. We validate the performance of our Global Pointer by comparing it with its alternatives:
	
	\begin{itemize}
		\item\textbf{Bert-CRF}. A baseline for entity extraction task that incorporates  pre-trained language model BERT \cite{devlin2018bert} and the additional Conditional Random Field (CRF) layer \cite{lafferty2001conditional}.
		\item \textbf{CopyRE} \cite{zeng2018extracting}.  An end-to-end model based on sequence-to-sequence learning with copy mechanism, which introduced to jointly extract  relation and entity from sentences.
		\item \textbf{GraphRel} \cite{fu-etal-2019-graphrel}. An end-to-end relation extraction model built upon the graph convolutional networks to jointly learn named entities and their corresponding relations.
		\item \textbf{CasRel} \cite{wei-etal-2020-novel}. A cascade binary tagging-based framework introduced to treat relations as functions mapping subjects to objects in a sentence to alleviate the overlapping problem in relation extraction.
		\item\textbf{ PFN} \cite{YanZhang-351}. A partition filter network-based approach introduced to model two-way interaction between  entity and relation extraction tasks. 
		
	\end{itemize}
	Moreover, we also compare to the baselines that achieve competitive performance, including Multi-head \cite{bekoulis-etal-2018-adversarial},  Multi-head + AT \cite{bekoulis2018joint}, Rel-Metric \cite{tran2019neural}, SpERT \cite{eberts2019span}.
	
	\begin{table}
		\centering
		\begin{tabular}{ l c c c c c}  
			\hline
			\textbf{Method}     & The People's daily & CLUENER & CMeEE &CONLL04 & Genia\\
			\hline
			Bert-CRF		& 95.46 			& 78.70  		& 64.39 			& 85.46 		& 73.02 \\ 
			PFN	\cite{YanZhang-351}			& 94.00  			& 79.29  		& 63.68 			& 87.43 		& 74.31 \\ 
			Global Pointer 	& \textbf{95.51 }	& \textbf{79.44}&\textbf{ 65.98} 	& \textbf{88.57}&\textbf{74.64} \\ 
			\hline
			
		\end{tabular}
		\caption{Comparative evaluation on various benchmark dataset for flat and nested NER. The results represent the Macro-F1 scores averaged of five runs with different randomization. The Note that all the results are our implementations and best scores are highlighted in bold.}
		\label{tab:main_results}	
	\end{table}
	\subsection{Main results}
	We use the Dev set to select the best model and report the average of five runs on each dataset as shown in Table \ref{tab:main_results} from which we have made the following observations: (1) our proposed solution gives the best Macro-F1 scores compared to the baselines across all datasets; (2) our Global Pointer can significantly outperform  BERT-CRF with more challenging datasets. For example, Global Pointer can achieve even about 0.74 and 1.59 with  CLUENER CMeEE datasets, respectively, over  BERT-CRF. Due to the widely recognized challenge of these datasets,  the achieved improvements can be deemed very considerable. Moreover, the experimental results in Table \ref{tab:results} have shown that our proposed solution can achieve a competitive performance compared to the state-of-the-art baselines with less training and inference costs.

	Furthermore, we compared Global Pointer to its alternative Bert-CRF in terms of computational costs of both training and inference steps. The comparative results are reported in Table \ref{tab:speed}. As can be seen,  our Global Pointer is faster than CRF, especially, with large datasets, such as the People's daily and CMeEE.

	\subsection{ Relative Position \& Class Imbalance loss Evaluation }
	To illustrate the affect of  encoding the relative position information, we conduct an ablation study  on the CONLL04 dataset as follows.  We drop Non-ROPE encoding component of our Global Pointer and compare the performance as shown in Table \ref{tab:abilation}. As can be seen, the Macro-F1 scores drop even about 11.43\%, and thus suggests that a well-designed mechanism that leverages the relative position information can boost the performance on NER task. 
	Moreover, we validate the efficacy of the proposed class imbalance loss function as follows. We replace the proposed loss function with the binary cross-entropy (BCE). We observe that the performance of  Global Pointer with BCE drops in terms of precision and F1 scores and thus demonstrates the effectiveness of our proposed loss function.     
	\subsection{Reduce Parameters Evaluation}

	\begin{table}
		\centering
			\begin{tabular}{lc}  
				\hline 
				
				\hline 
				Method & F1 score\\
				\hline 
				\textbf{NYT} $\vartriangle$\\
				CopyRE \cite{zeng2018extracting}& 86.2 \\
				GraphRel \cite{fu-etal-2019-graphrel}& 89.2 \\
				CasRel \cite{wei-etal-2020-novel} $^\dag$& 93.5\\
				PFN \cite{YanZhang-351}$^\dag$& \textbf{95.8} \\
				\hline 
				Global Pointer $^\dag$ & 95.6\\
				\hline 
				
				\textbf{WebNLG} $\vartriangle$\\
				CopyRE \cite{zeng2018extracting} &82.1 \\
				GraphRel \cite{fu-etal-2019-graphrel} &91.9 \\
				CasRel \cite{wei-etal-2020-novel}$^\dag$ &95.5 \\
				PFN \cite{YanZhang-351}$^\dag$&\textbf{98.0}\\
				\hline 
				Global Pointer $^\dag$ & \textbf{98.0}\\
				\hline 
				\textbf{ADE} $\blacktriangle$\\
				Multi-head \cite{bekoulis-etal-2018-adversarial}  &86.4\\
				Multi-head + AT \cite{bekoulis2018joint} & 86.7 \\
				Rel-Metric (Tran and Kavuluru, 2019) & 87.1 \\
				SpERT \cite{eberts2019span}$^\dag$ &89.3\\
				
				
				PFN   \cite{YanZhang-351}$^\dag$& 89.6\\
				\hline 
				Global Pointer $^\dag$ &\textbf{90.1}\\
				\hline 
				
			\end{tabular}
		\caption{Comparative evaluation,  $^\dag$,$^\ddag$ and $^\S$	denotes the use of BERT, ALBERT and SCIBERT \cite{devlin2018bert,lan2019albert,beltagy-etal-2019-scibert} pre-trained embedding. $\vartriangle$ and $\blacktriangle$ denotes
			the use of micro-F1 and macro-F1 score.}
		\label{tab:results}	
	\end{table}
	\begin{table}
		\centering
		\begin{tabular}{ l c c c cc }  
			\hline
			\multirow{2}{*}{Dataset} &\multicolumn{2}{c}{Training Speed}&&\multicolumn{2}{c}{Inference Speed}\\ \cline{2-3} \cline{5-6}
			&BERT-CRF&Global Pointer&&BERT-CRF&Global Pointer\\
			\hline
			The People's daily	& 1x & 1.56x && 1x  & 1.11x \\ 
			CLUENER				& 1x & 1.22x && 1x  & 1x \\ 
			CMeEE 				& 1x & 1.52x && 1x  & 1.13x\\ 			
			\hline
		\end{tabular}
		\caption{Comparative evaluation in terms of computational cost between the proposed  Global Pointer and BERT-CRF }
		\label{tab:speed}	
	\end{table}
	\begin{table}
	\centering
	\begin{tabular}{ l c c c c }  
		\hline	
		\textbf{Dataset}  & Global Pointer & Efficient Global Pointer \\
		\hline
		The People's daily	& \textbf{95.51} & 95.36   \\ 
		CLUENER	& 79.44  & \textbf{80.04} \\ 
		CMeEE 	& 65.98 & \textbf{66.54}\\ 	
		\hline		
	\end{tabular}
	\caption{Comparison of the Efficient Global Pointer with the original Global Pointer in F1 score. Best scores are highlighted in bold.}
	\label{tab:EGP_results}	
\end{table}
	\begin{table}
		\centering
		\begin{tabular}{ccccccccc}  
			\hline
			\multirow{2}{*}{Category}&\multirow{2}{*}{Number of sentences}&\multicolumn{3}{c}{Global Pointer}&&\multicolumn{3}{c}{PFN}\\ \cline{3-5}\cline{7-9}
			&  & P & R & F1 && P & R & F1 \\
			\hline
			L-1	& 120 	&\textbf{ 94.24} & 90.82 & \textbf{92.50} &	&91.30 & \textbf{91.30} &91.30 \\ 
			L-2	& 161  	& \textbf{86.42} &\textbf{ 85.87} & \textbf{86.15} &	& 85.04 & 85.71 & 85.38 \\ 
			L-3 & 7 	&\textbf{ 82.05} & \textbf{91.43} & \textbf{86.49} &	& 72.09 & 88.57 & 79.49\\
			
			D-1 & 10 & \textbf{100.0} &\textbf{ 90.0 }& \textbf{94.74 }&	&\textbf{100.0} &\textbf{ 90.0} &\textbf{94.74}\\
			D-2 & 172 & \textbf{87.23} &87.23 &\textbf{ 87.23} &	& 84.07 &\textbf{87.69} & 85.84\\
			D-3	& 120 & \textbf{94.24} & 90.82 & \textbf{92.50} &	&91.30 & \textbf{91.30} & 91.30\\ 	
			\hline			
		\end{tabular}
		\caption{Comparative evaluation of Global Pointer and PFN in terms of entity length and entity density. Best scores are highlighted in bold.}
		\label{tab:GP_SN_results}	
	\end{table}
	
	\begin{table}
		\centering
		\begin{tabular}{ l c c c }  
			\hline
			\textbf{Ablations}  & Precision & Recall & F1 \\
			\hline
			Non ROPE		& 77.14 & 77.14 & 77.14   \\ 
			BCE loss		& 88.72  & \textbf{88.22} & 88.48 \\ 
			Global Pointer 	& \textbf{89.19} & 87.95 & \textbf{88.57}\\ 			
			\hline
		\end{tabular}
		\caption{The comparative evaluation of relative position information and  class imbalance loss on CONLL04 dataset.}
		\label{tab:abilation}	
	\end{table}

	In Section \ref{sec:reduce_param}, we introduce a new variant of the proposed solution, namely Efficient Global Pointer, which can perform under less parameters settings. We conduct empirical experiments on the people's daily, CLUENER and CMeEE datasets to evaluate the performance of both variants. The comparative results are shown in Table \ref{tab:EGP_results} from which we have made the following observations.
	(1) Overall, Efficient Global Pointer can mostly give the best F1 scores. (2)  Despite the limited number of parameters, Efficient Global Pointer can still be competitive on the easy dataset, e.g., People's daily dataset. (3) CLUENER and CMeEE were annotated with 10 and 9 entity types, respectively, which are widely recognized as more challenging datasets; however, Efficient Global Pointer with less parameters can still perform better than its alternative with all parameters. The performance is expected as the number of parameters increases with each entity type leading to an overfiting problem. In brief, the experimental results suggest that a carefully-designed mechanism to reduce the number of parameters can enhance the performance of NER. 
	
	
	%
	
	\subsection{Empirical Analysis}

	In the section, we perform in-depth analysis in terms of entity length and entity density. Specifically, we conducted relevant experiments on CONLL04 dataset to evaluate the performance of Global Pointer and PFN \cite{YanZhang-351}. First, we map the sentences into three groups according to their length: $L<3$, $3=<L<6$, and $L>=6$, denoted as  $L_1$, $L_2$ and $L_3$, respectively. 
	Second, we categorized the sentences  according to their density: dense<=0.1, 0.1 < dense < = 0.3, dense > 0.3, denoted as  $D_1$, $D_2$ and $D_3$. Note that we use the ratio of the number of entity words to the total number of text words as the index of entity density.
	
	
	The comparative evaluation is depicted in Table \ref{tab:GP_SN_results}. We observe that when the entity length exceeds the half (e.g., 6), Global Pointer can achieve even about 7\% improvements higher than PFN in terms of F1 score. These improvements demonstrate the importance of relative position information in the large number of entities recognition. In addition, we also observe that when the density of entities in the text is at the middle level, both models give the worse scores.  However, as can be seen, Global Pointer performs better in most scenarios.

	\section{Conclusions} \label{sec:conclusion}
	
	In this paper, we presented a novel solution to address span-based NER framework, namely Global Pointer (GP), by leveraging the relative positions through a multiplicative attention mechanism. 
	GP is designed of two modules that aim to identify the head and the tail of a given entity to enable the inconsistency between the training and inference processes. Moreover, GP contributed with a novel  loss function to address the imbalance label problem. To reduce the training cost, we introduced a new variant of GP based on  approximate method to reduce the training parameters.
	We extensively evaluated GP on various benchmark datasets. Our extensive experiments demonstrate that GP can outperform the existing solution. Moreover, the experimental results show the efficacy of the introduced loss function compared to softmax and entropy alternatives.  
	\bibliographystyle{unsrtnat}
	\bibliography{references}  
	
\end{document}